\documentclass[letterpaper]{article} 
\usepackage{aaai25}  
\usepackage{times}  
\usepackage{helvet}  
\usepackage{courier}  
\usepackage[hyphens]{url}  
\usepackage{graphicx} 
\usepackage{natbib}  
\usepackage{caption} 
\usepackage{siunitx}

\usepackage[T1]{fontenc}
\usepackage[utf8]{inputenc}
\usepackage{mathtools}

\usepackage{amssymb,mathrsfs}
\usepackage{amsthm}
\usepackage{bm}
\usepackage{scalerel}
\usepackage{nicefrac}
\usepackage{microtype} 
\usepackage[shortlabels]{enumitem}
\usepackage{graphicx}
\usepackage{epstopdf}
\DeclareGraphicsExtensions{.eps,.png,.jpg,.pdf}

\usepackage{colortbl}
\usepackage{booktabs}
\usepackage{multirow}
\usepackage{colortbl,xcolor}
\usepackage{xparse,xstring}
\usepackage{calc}
\usepackage{etoolbox}

\usepackage{array}
\newcolumntype{L}[1]{>{\raggedright\let\newline\\\arraybackslash\hspace{0pt}}m{#1}}
\newcolumntype{C}[1]{>{\centering\let\newline\\\arraybackslash\hspace{0pt}}m{#1}}
\newcolumntype{R}[1]{>{\raggedleft\let\newline\\\arraybackslash\hspace{0pt}}m{#1}}

\makeatletter
\let\MYcaption\@makecaption
\makeatother
\usepackage[font=footnotesize]{subcaption}
\makeatletter
\let\@makecaption\MYcaption
\makeatother

\usepackage{glossaries}
\makeatletter
\sfcode`\.1006

\let\oldgls\gls
\let\oldglspl\glspl

\newcommand\fussy@ifnextchar[3]{%
	\let\reserved@d=#1%
	\def\reserved@a{#2}%
	\def\reserved@b{#3}%
	\futurelet\@let@token\fussy@ifnch}
\def\fussy@ifnch{%
	\ifx\@let@token\reserved@d
		\let\reserved@c\reserved@a
	\else
		\let\reserved@c\reserved@b
	\fi
	\reserved@c}

\renewcommand{\gls}[1]{%
\oldgls{#1}\fussy@ifnextchar.{\@checkperiod}{\@}}
\renewcommand{\glspl}[1]{%
\oldglspl{#1}\fussy@ifnextchar.{\@checkperiod}{\@}}

\newcommand{\@checkperiod}[1]{%
	\ifnum\sfcode`\.=\spacefactor\else#1\fi
}

\robustify{\gls}
\robustify{\glspl}
\makeatother

\newacronym{wrt}{w.r.t.}{with respect to}
\newacronym{RHS}{R.H.S.}{right-hand side}
\newacronym{LHS}{L.H.S.}{left-hand side}
\newacronym{iid}{i.i.d.}{independent and identically distributed}
\newacronym{SOTA}{SOTA}{state-of-the-art}

\usepackage{float}

\usepackage[capitalize]{cleveref}
\crefname{equation}{}{}
\Crefname{equation}{}{}
\crefname{claim}{claim}{claims}
\crefname{step}{step}{steps}
\crefname{line}{line}{lines}
\crefname{condition}{condition}{conditions}
\crefname{dmath}{}{}
\crefname{dseries}{}{}
\crefname{dgroup}{}{}

\crefname{Problem}{Problem}{Problems}
\crefformat{Problem}{Problem~#2#1#3}
\crefrangeformat{Problem}{Problems~#3#1#4 to~#5#2#6}

\crefname{Theorem}{Theorem}{Theorems}
\crefname{Corollary}{Corollary}{Corollaries}
\crefname{Proposition}{Proposition}{Propositions}
\crefname{Lemma}{Lemma}{Lemmas}
\crefname{Definition}{Definition}{Definitions}
\crefname{Example}{Example}{Examples}
\crefname{Assumption}{Assumption}{Assumptions}
\crefname{Remark}{Remark}{Remarks}
\crefname{Rem}{Remark}{Remarks}
\crefname{remarks}{Remarks}{Remarks}
\crefname{Appendix}{Appendix}{Appendices}
\crefname{Supplement}{Supplement}{Supplements}
\crefname{Exercise}{Exercise}{Exercises}
\crefname{Theorem_A}{Theorem}{Theorems}
\crefname{Corollary_A}{Corollary}{Corollaries}
\crefname{Proposition_A}{Proposition}{Propositions}
\crefname{Lemma_A}{Lemma}{Lemmas}
\crefname{algorithm}{Algorithm}{Algorithms}
\crefname{Definition_A}{Definition}{Definitions}

\usepackage{crossreftools}

\usepackage{algorithm}
\usepackage{algpseudocode}

\ifx\loadbreqn\undefined
	\relax
\else
	\usepackage{breqn}
\fi

\makeatletter
\def\cleartheorem#1{%
    \expandafter\let\csname#1\endcsname\relax
    \expandafter\let\csname c@#1\endcsname\relax
}
\def\clearthms#1{ \@for\tname:=#1\do{\cleartheorem\tname} }
\makeatother

\ifx\renewtheorem\undefined
	\ifx\useTheoremCounter\undefined
		\newtheorem{Theorem}{Theorem}
		\newtheorem{Corollary}{Corollary}
		\newtheorem{Proposition}{Proposition}
		
	\else

	\fi

\fi

\theoremstyle{remark}

\theoremstyle{plain}

\newcommand{\qednew}{\nobreak \ifvmode \relax \else
		\ifdim\lastskip<1.5em \hskip-\lastskip
			\hskip1.5em plus0em minus0.5em \fi \nobreak
		\vrule height0.75em width0.5em depth0.25em\fi}



\newcommand{\nn}{\nonumber\\ }

\NewDocumentCommand{\movedownsub}{e{^_}}{%
	\IfNoValueTF{#1}{%
		\IfNoValueF{#2}{^{}}
	}{%
		^{#1}
	}%
	\IfNoValueF{#2}{_{#2}}
}

\let\latexchi\chi
\RenewDocumentCommand{\chi}{}{\latexchi\movedownsub}

\newcommand{\Real}{\mathbb{R}}

\newcommand{\bz}{\mathbf{z}}

\DeclareSymbolFont{bsfletters}{OT1}{cmss}{bx}{n}
\DeclareSymbolFont{ssfletters}{OT1}{cmss}{m}{n}
\DeclareMathSymbol{\bsfGamma}{0}{bsfletters}{'000}
\DeclareMathSymbol{\ssfGamma}{0}{ssfletters}{'000}
\DeclareMathSymbol{\bsfDelta}{0}{bsfletters}{'001}
\DeclareMathSymbol{\ssfDelta}{0}{ssfletters}{'001}
\DeclareMathSymbol{\bsfTheta}{0}{bsfletters}{'002}
\DeclareMathSymbol{\ssfTheta}{0}{ssfletters}{'002}
\DeclareMathSymbol{\bsfLambda}{0}{bsfletters}{'003}
\DeclareMathSymbol{\ssfLambda}{0}{ssfletters}{'003}
\DeclareMathSymbol{\bsfXi}{0}{bsfletters}{'004}
\DeclareMathSymbol{\ssfXi}{0}{ssfletters}{'004}
\DeclareMathSymbol{\bsfPi}{0}{bsfletters}{'005}
\DeclareMathSymbol{\ssfPi}{0}{ssfletters}{'005}
\DeclareMathSymbol{\bsfSigma}{0}{bsfletters}{'006}
\DeclareMathSymbol{\ssfSigma}{0}{ssfletters}{'006}
\DeclareMathSymbol{\bsfUpsilon}{0}{bsfletters}{'007}
\DeclareMathSymbol{\ssfUpsilon}{0}{ssfletters}{'007}
\DeclareMathSymbol{\bsfPhi}{0}{bsfletters}{'010}
\DeclareMathSymbol{\ssfPhi}{0}{ssfletters}{'010}
\DeclareMathSymbol{\bsfPsi}{0}{bsfletters}{'011}
\DeclareMathSymbol{\ssfPsi}{0}{ssfletters}{'011}
\DeclareMathSymbol{\bsfOmega}{0}{bsfletters}{'012}
\DeclareMathSymbol{\ssfOmega}{0}{ssfletters}{'012}

\newcommand{\btheta}{\bm{\theta}}

\makeatletter
\newcommand*\rel@kern[1]{\kern#1\dimexpr\macc@kerna}
\newcommand*\widebar[1]{%
  \begingroup
  \def\mathaccent##1##2{%
    \rel@kern{0.8}%
    \overline{\rel@kern{-0.8}\macc@nucleus\rel@kern{0.2}}%
    \rel@kern{-0.2}%
  }%
  \macc@depth\@ne
  \let\math@bgroup\@empty \let\math@egroup\macc@set@skewchar
  \mathsurround\z@ \frozen@everymath{\mathgroup\macc@group\relax}%
  \macc@set@skewchar\relax
  \let\mathaccentV\macc@nested@a
  \macc@nested@a\relax111{#1}%
  \endgroup
}
\makeatother

\DeclareMathOperator{\var}{var}

\DeclareMathOperator{\cov}{cov}

\newcommand{\ifbcdot}[1]{\ifblank{#1}{\cdot}{#1}}

\DeclarePairedDelimiterX\abs[1]{\lvert}{\rvert}{\ifbcdot{#1}}
\DeclarePairedDelimiterX\parens[1]{(}{)}{\ifbcdot{#1}}
\DeclarePairedDelimiterX\brk[1]{[}{]}{\ifbcdot{#1}}
\DeclarePairedDelimiterX\braces[1]{\{}{\}}{\ifbcdot{#1}}
\DeclarePairedDelimiterX\angles[1]{\langle}{\rangle}{\ifblank{#1}{\cdot,\cdot}{#1}}
\DeclarePairedDelimiterX\ip[2]{\langle}{\rangle}{\ifbcdot{#1},\ifbcdot{#2}}
\DeclarePairedDelimiterX\norm[1]{\lVert}{\rVert}{\ifbcdot{#1}}
\DeclarePairedDelimiterX\ceil[1]{\lceil}{\rceil}{\ifbcdot{#1}}
\DeclarePairedDelimiterX\floor[1]{\lfloor}{\rfloor}{\ifbcdot{#1}}

\DeclareFontFamily{U}{matha}{\hyphenchar\font45}
\DeclareFontShape{U}{matha}{m}{n}{
      <5> <6> <7> <8> <9> <10> gen * matha
      <10.95> matha10 <12> <14.4> <17.28> <20.74> <24.88> matha12
      }{}
\DeclareSymbolFont{matha}{U}{matha}{m}{n}
\DeclareFontSubstitution{U}{matha}{m}{n}

\DeclareFontFamily{U}{mathx}{\hyphenchar\font45}
\DeclareFontShape{U}{mathx}{m}{n}{
      <5> <6> <7> <8> <9> <10>
      <10.95> <12> <14.4> <17.28> <20.74> <24.88>
      mathx10
      }{}
\DeclareSymbolFont{mathx}{U}{mathx}{m}{n}
\DeclareFontSubstitution{U}{mathx}{m}{n}

\DeclareMathDelimiter{\vvvert}{0}{matha}{"7E}{mathx}{"17}
\DeclarePairedDelimiterX\vertiii[1]{\vvvert}{\vvvert}{\ifbcdot{#1}}

\DeclarePairedDelimiterXPP\trace[1]{\operatorname{Tr}}{(}{)}{}{\ifbcdot{#1}} 
\DeclarePairedDelimiterXPP\col[1]{\operatorname{col}}{\{}{\}}{}{\ifbcdot{#1}} 
\DeclarePairedDelimiterXPP\row[1]{\operatorname{row}}{\{}{\}}{}{\ifbcdot{#1}} 
\DeclarePairedDelimiterXPP\erf[1]{\operatorname{erf}}{(}{)}{}{\ifbcdot{#1}}
\DeclarePairedDelimiterXPP\erfc[1]{\operatorname{erfc}}{(}{)}{}{\ifbcdot{#1}}
\DeclarePairedDelimiterXPP\KLD[2]{D}{(}{)}{}{\ifbcdot{#1}\, \delimsize\|\, \ifbcdot{#2}} 
\DeclarePairedDelimiterXPP\op[2]{\operatorname{#1}}{(}{)}{}{#2} 

\newcommand{\ud}{\,\mathrm{d}} 

\DeclarePairedDelimiterXPP\indicate[1]{{\bf 1}}{\{}{\}}{}{\ifbcdot{#1}}

\NewDocumentCommand\ofrac{s m}{%
	\IfBooleanTF#1%
	{\dfrac{1}{#2}}%
	{\frac{1}{#2}}%
}
\NewDocumentCommand\ddfrac{s m m}{%
	\IfBooleanTF#1%
	{\dfrac{\mathrm{d} {#2}}{\mathrm{d} {#3}}}%
	{\frac{\mathrm{d} {#2}}{\mathrm{d} {#3}}}%
}
\NewDocumentCommand\ppfrac{s m m}{%
	\IfBooleanTF#1%
	{\dfrac{\partial {#2}}{\partial {#3}}}%
	{\frac{\partial {#2}}{\partial {#3}}}%
}

\providecommand\given{}

\DeclarePairedDelimiterX\Set[2]\{\}{%
\renewcommand\given{\SetSymbol[\delimsize]{#1}}
#2
}
\DeclarePairedDelimiterX\Setc[1]\{\}{%
\renewcommand\given{\SetSymbol{:}}
#1
}

\NewDocumentCommand\set{s o m}{%
	\IfBooleanTF#1%
	{\IfValueTF{#2}{\Set*{#2}{#3}}{\Setc*{#3}}}%
	{\IfValueTF{#2}{\Set{#2}{#3}}{\Setc{#3}}}%
}

\NewDocumentCommand{\evalat}{ s O{\big} m e{_^} }{%
\IfBooleanTF{#1}%
{\left. #3 \right|}{#3#2|}%
\IfValueT{#4}{_{#4}}%
\IfValueT{#5}{^{#5}}%
}

\providecommand\given{}
\DeclarePairedDelimiterXPP\cprob[1]{}(){}{
\renewcommand\given{\nonscript\,\delimsize\vert\allowbreak\nonscript\,\mathopen{}}%
#1%
}
\DeclarePairedDelimiterXPP\cexp[1]{}[]{}{
\renewcommand\given{\nonscript\,\delimsize\vert\allowbreak\nonscript\,\mathopen{}}%
#1%
}

\DeclareDocumentCommand \P { s e{_^} d() g } {%
	\mathbb{P}%
	\IfBooleanTF{#1}%
		{
			\IfValueT{#2}{_{#2}}%
			\IfValueT{#3}{^{#3}}%
			\IfValueTF{#5}{\cprob{#4 \given #5}}{\IfValueT{#4}{\cprob{#4}}}%
		}%
		{
			\IfValueT{#2}{_{#2}}%
			\IfValueT{#3}{^{#3}}%
			\IfValueTF{#5}{\cprob*{#4 \given #5}}{\IfValueT{#4}{\cprob*{#4}}}%
		}%
}

\DeclareDocumentCommand \E { s e{_^} o g } {%
	\mathbb{E}%
	\IfBooleanTF{#1}%
		{
			\IfValueT{#2}{_{#2}}%
			\IfValueT{#3}{^{#3}}%
			\IfValueTF{#5}{\cexp{#4 \given #5}}{\IfValueT{#4}{\cexp{#4}}}%
		}%
		{
			\IfValueT{#2}{_{#2}}%
			\IfValueT{#3}{^{#3}}%
			\IfValueTF{#5}{\cexp*{#4 \given #5}}{\IfValueT{#4}{\cexp*{#4}}}%
		}%
}

\DeclareDocumentCommand \Var { s e{_^} d() g } {%
	\var%
	\IfBooleanTF{#1}%
		{
			\IfValueT{#2}{_{#2}}%
			\IfValueT{#3}{^{#3}}%
			\IfValueTF{#5}{\cprob{#4 \given #5}}{\IfValueT{#4}{\cprob{#4}}}%
		}%
		{
			\IfValueT{#2}{_{#2}}%
			\IfValueT{#3}{^{#3}}%
			\IfValueTF{#5}{\cprob*{#4 \given #5}}{\IfValueT{#4}{\cprob*{#4}}}%
		}%
}

\DeclareDocumentCommand \Cov { s e{_^} d() g } {%
	\cov%
	\IfBooleanTF{#1}%
		{
			\IfValueT{#2}{_{#2}}%
			\IfValueT{#3}{^{#3}}%
			\IfValueTF{#5}{\cprob{#4 \given #5}}{\IfValueT{#4}{\cprob{#4}}}%
		}%
		{
			\IfValueT{#2}{_{#2}}%
			\IfValueT{#3}{^{#3}}%
			\IfValueTF{#5}{\cprob*{#4 \given #5}}{\IfValueT{#4}{\cprob*{#4}}}%
		}%
}

\ExplSyntaxOn
\NewDocumentCommand \dist {m o o} {%
\mathrm{#1}\left(%
	\IfValueT{#3}{%
		\tl_if_blank:nTF{ #3 }{\cdot\, \middle|\, }{#3\, \middle|\, }%
	}
	\IfValueT{#2}{#2}%
\right)%
}
\ExplSyntaxOff

\NewDocumentCommand {\cbrace} {t+ D[]{black} D(){\widthof{#5}} m m } {%
	\begingroup%
		\color{#2}
		\IfBooleanTF{#1}{%
			\overbrace{#4}^%
		}{
			\underbrace{#4}_%
		}%
		{\parbox[c]{#3}{\centering\footnotesize{#5}}}%
	\endgroup%
}

\let\oldforall\forall
\renewcommand{\forall}{\oldforall \, }

\let\oldexist\exists
\renewcommand{\exists}{\oldexist \, }

\makeatletter

\newcommand{\rankcolor}[2]{%
	\expandafter\renewcommand\csname #1\endcsname[1]{%
		\ifblank{##1}{%
			{\color{#2} \textbf{#2}}%
		}{%
			\ifmmode
				\textcolor{#2}{\bm{##1}}%
			\else%
				{\color{#2} \textbf{##1}}%
			\fi	
		}%
	}
}

\providecommand{\first}{}
\providecommand{\second}{}

\rankcolor{first}{red}
\rankcolor{second}{blue}
\rankcolor{third}{cyan}
\makeatother

\graphicspath{{./Figures/}{./figures/}}
\DeclareDocumentCommand{\includeCroppedPdf}{ o O{./Figures/} m }{
	\IfFileExists{#2#3-crop.pdf}{}{%
		\immediate\write18{pdfcrop #2#3.pdf #2#3-crop.pdf}}%
	\includegraphics[#1]{#2#3-crop.pdf}
}

\makeatletter
\newcommand*{\addFileDependency}[1]{
  \typeout{(#1)}
  \@addtofilelist{#1}
  \IfFileExists{#1}{}{\typeout{No file #1.}}
}
\makeatother

\definecolor{gray90}{gray}{0.9}
\def\colorlist{red,blue,brown,cyan,darkgray,gray,lightgray,green,lime,magenta,olive,orange,pink,purple,teal,violet,white,yellow}

\makeatletter
\def\startcomment{[}
\ifx\nohighlights\undefined
	\newcommand{\createcolor}[1]{%
			\expandafter\newcommand\csname #1\endcsname[1]{{\color{#1} ##1}}%
	}
	\newcommand{\msout}[1]{\text{\color{green} \sout{\ensuremath{#1}}}}
	\newcommand{\del}[1]{{\color{green}\ifmmode \msout{#1}\else\sout{#1}\fi}}
\else
	\newcommand{\createcolor}[1]{%
			\expandafter\newcommand\csname #1\endcsname[1]{%
				\noexpandarg%
				\StrChar{##1}{1}[\firstletter]%
				\if\firstletter\startcomment%
					\relax
				\else%
					##1
				\fi
			}%
	}
	\newcommand{\msout}[1]{}
	\newcommand{\del}[1]{}
\fi

\def\@tempa#1,{%
    \ifx\relax#1\relax\else
        \createcolor{#1}%
        \expandafter\@tempa
    \fi
}
\expandafter\@tempa\colorlist,\relax,
\makeatother

\newcommand{\hhide}[1]{}

\ifx\diagnoselabel\undefined
	\relax
\else
	\makeatletter
	\def\@testdef #1#2#3{%
		\def\reserved@a{#3}\expandafter \ifx \csname #1@#2\endcsname
			\reserved@a  \else
			\typeout{^^Jlabel #2 changed:^^J%
				\meaning\reserved@a^^J%
				\expandafter\meaning\csname #1@#2\endcsname^^J}%
			\@tempswatrue \fi}
	\makeatother
\fi

\newcommand{\tb}[1]{\textbf{#1}}

\renewcommand{\first}{\textbf}
\renewcommand{\second}{\underline}

\usepackage{multirow}
\usepackage{booktabs}
\usepackage{pifont}
\usepackage{bm}
\usepackage{enumitem}
\usepackage{newfloat}
\usepackage{listings}
\DeclareCaptionStyle{ruled}{labelfont=normalfont,labelsep=colon,strut=off} 
\floatstyle{ruled}
\newfloat{listing}{tb}{lst}{}
\floatname{listing}{Listing}
\usepackage{makecell}

\makeatletter
\renewcommand*{\@fnsymbol}[1]{\ensuremath{\ifcase#1\or \dagger \or * \or \ddagger\or
   \mathsection\or \mathparagraph\or \|\or **\or \dagger\dagger
   \or \ddagger\ddagger \else\@ctrerr\fi}}
\makeatother
\title{Efficient Training of Neural Fractional-Order Differential Equation\\ via Adjoint Backpropagation}
\author{
    Qiyu Kang\textsuperscript{\rm 1}, 
    Xuhao Li\thanks{Corresponding author (lixh@ahu.edu.cn).}\textsuperscript{\rm 2}, 
    Kai Zhao\textsuperscript{\rm 3}, 
    Wenjun Cui\textsuperscript{\rm 4},
    Yanan Zhao\textsuperscript{\rm 3},
    Weihua Deng\textsuperscript{\rm 5},
    Wee Peng Tay\textsuperscript{\rm 3} 
}
\affiliations{
    \textsuperscript{\rm 1} University of Science and Technology of China \\
      \textsuperscript{\rm 2} Anhui University \\
  \textsuperscript{\rm 3} Nanyang Technological University \\
      \textsuperscript{\rm 4} Beijing Jiaotong University \\
      \textsuperscript{\rm 5} Lanzhou University\\

}

\usepackage{textgreek} 

\usepackage{bibentry}

\begin{document}

\maketitle

\begin{abstract}
Fractional-order differential equations (FDEs) enhance traditional differential equations by extending the order of differential operators from integers to real numbers, offering greater flexibility in modeling complex dynamical systems with nonlocal characteristics.
Recent progress at the intersection of FDEs and deep learning has catalyzed a new wave of innovative models, demonstrating the potential to address challenges such as graph representation learning.
However, training neural FDEs has primarily relied on direct differentiation through forward-pass operations in FDE numerical solvers, leading to increased memory usage and computational complexity, particularly in large-scale applications. To address these challenges, we propose a scalable adjoint backpropagation method for training neural FDEs by solving an augmented FDE backward in time, which substantially reduces memory requirements. 
This approach provides a practical neural FDE toolbox and holds considerable promise for diverse applications. 
We demonstrate the effectiveness of our method in several tasks, achieving performance comparable to baseline models while significantly reducing computational overhead.  
\end{abstract}

\begin{links} 
\link{Code}{https://github.com/kangqiyu/torchfde}
\end{links}

\section{Introduction}

Fractional calculus is a mathematical generalization of integer-order integration and differentiation, enabling the modeling of complex processes in physical systems beyond what traditional calculus can achieve. It finds applications across multiple disciplines, illustrating its versatility.
For example, it characterizes viscoelastic materials \cite{coleman1961foundations}, models population dynamics \cite{almeida2016modeling}, enhances control systems \cite{podlubny1994fractional}, improves signal processing \cite{machado2011recent}, supports financial modeling \cite{scalas2000fractional}, and describes porous and fractal structures \cite{nigmatullin1986realization,mandelbrot1982fractal,ionescu2017role}.
Within these varied contexts, fractional-order differential equations (FDEs) serve as an enriched extension of traditional integer-order differential equations, providing a way to incorporate a continuum of past states into the present state. 
This generalization enables the capture of memory and non-locality effects inherent in various physical and engineering processes.

In the realm of deep learning, traditional neural Ordinary Differential Equations (ODEs) predominantly rely on integer-order differential equations, which can be conceptualized as continuous residual layers. \cite{weinan2017proposal,chen2018neural,kidger2021hey}. 
These models have been widely applied in areas such as content generation \cite{yang2023diffusion, song2020score}, adversarial robustness \cite{kang2021stable, yan2019robustness}, and physics modeling \cite{ji2021stiff,raissi2019physics,lai2021structural}. 
Despite their widespread success, integer-order ODEs struggle to effectively capture the complex memory-dependent characteristics of systems due to the inherent limitations of integer-order operators \cite{podlubny1994fractional}.
The integration of fractional calculus with deep learning has recently garnered interest from researchers in addressing the shortcomings of integer-order models.
Innovations in this area include the application of fractional derivatives for optimizing parameters in neural networks \cite{liu2022regularized}, moving away from traditional gradient optimization methods like SGD or Adam \cite{kingma2014adam}. Furthermore, \cite{antil2020fractional} has demonstrated that incorporating fractional calculus with its L1 approximation can enable networks to handle non-smooth data while addressing the vanishing gradient problem.
Notably, in the field of physics-informed machine learning, the development of fractional physics-informed neural networks (fPINNs) \cite{pang2019fpinns}  highlights how these networks incorporate FDEs to embed physics principles, setting a new direction in the literature \cite{guo2022monte, javadi2023solving, wang2022fractional}.
Additionally, the application of neural FDEs for updating graph hidden features has been shown to improve model performance, alleviate oversmoothing, and strengthen adversarial defense \cite{KanZhaDin:C24frond, ZhaKanSon:C24robustfrond, ZhaKanJi:C24, CuiKanLi:C25}.

In contemporary research, training integer-order neural ODEs leverages reverse-mode differentiation by solving an augmented ODE backward in time, which is memory efficient \cite{chen2018neural}. 
In contrast, training neural FDEs still relies on the basic automatic differentiation technique from leading platforms such as TensorFlow \cite{abadi2016tensorflow} and PyTorch \cite{paszke2019pytorch}. 
This approach involves backpropagation through forward-pass operations in FDE numerical solvers, requiring multiple iterations and are computationally intensive, which poses significant challenges for efficient training. 
To address this challenge, we introduce a scalable method that facilitates backpropagation by solving an augmented FDE backward in time. 
This method enables seamless end-to-end training of FDE-based modules within larger models and uses less memory.
We validate our approach in practical tasks, demonstrating that it achieves performance on par with baseline models while significantly reducing computational memory demands during training.

\tb{Main contributions.} 
Our key contributions are summarized as follows:
\begin{itemize}
\item We propose a novel and efficient method for training neural FDEs by solving an augmented FDE backward in time. This approach not only reduces computational training memory usage but also facilitates the efficient training of FDE components within larger models.
\item We develop a practical neural FDE toolbox that has the potential for diverse applications. Our method has been successfully applied to practical graph representation learning tasks, achieving performance comparable to established baseline models while requiring less computational memory for training.
\end{itemize}

The rest of the paper is organized as follows: 
\cref{sec.related} offers a review of the literature related to the use of differential equations in machine learning.
\cref{sec.prelim} briefly covers the mathematical foundations of fractional calculus to assist readers who are unfamiliar with it. 
\cref{sec.framework} outlines our proposed neural FDE parameter backpropagation, detailing the rationale behind its design.
\cref{sec.exp} describes the experimental procedures and presents the results.
We summarize and conclude the paper in \cref{sec.conclusion}.

\section{Related Work}\label{sec.related}
In this section, we provide a literature review of neural differential equations, fractional calculus, and their applications in machine learning. 

\subsection{Differential Equations and Machine Learning}
The combination of differential equations with machine learning represents a significant leap forward in tackling complex problems across various domains \cite{raissi2019physics,weinan2017proposal,chen2018neural}. 
This innovative approach fuses accurate dynamical system modeling via differential equations with the high expressivity of neural networks. 
Among its many applications, particularly notable is its use in predictive modeling and simulation of physical systems. 
For instance, neural ODE \cite{chen2018neural} represents a significant advancement where the traditional layers of a neural network are replaced with continuous-depth models, enabling the network to learn complex dynamics with potentially fewer parameters and enhanced interpretability compared to standard deep learning models. 
Additionally, this integration may enhance neural network performance \cite{dupont2019augmented}, stabilize gradients \cite{haber2017stable, gravina2022anti}, and increase neural network robustness \cite{yan2019robustness, kang2021stable,WanKanShe:C23}.
In computational fluid dynamics, machine learning models integrated with differential equations help in accelerating simulations without compromising on accuracy \cite{miyanawala2017efficient}. 
Moreover, the integration of these disciplines also facilitates the development of data-driven discovery of differential equations. Techniques such as sparse identification of nonlinear dynamical systems \cite{brunton2016discovering,wang2023scientific} allow scientists to discover the underlying differential equations from experimental data, essentially learning the laws of physics governing a particular system. 
Overall, the synergy between differential equations and machine learning not only enhances computational efficiency and model accuracy but also opens up new research questions and methodologies. It promotes an active exchange of ideas and leads to innovations that could be beneficial across various domains.

\subsection{Fractional Calculus and Its Applications}
The field of fractional calculus, extending the traditional definitions of calculus to non-integer orders, has evolved significantly, offering new perspectives and tools for solving various scientific and engineering problems. This approach is pivotal for processes exhibiting anomalous or non-local properties that classical integer-order methods fail to capture. In engineering, fractional calculus enhances system control, achieving greater stability and performance with fractional-order controllers \cite{podlubny1994fractional}. It also aids in modeling electrical circuits and materials more accurately \cite{kaczorek2015fractional}. In physics, it provides a framework for describing anomalous diffusion in media like geological formations, capturing dynamics unexplained by classical theories \cite{diaz2022time,sornette2006critical}.
In finance, fractional derivatives model the heavy tails and memory effects of financial time series, leading to more precise risk management tools \cite{scalas2000fractional}. Meanwhile, in medicine and biology, it models phenomena such as blood flow in aneurysms and epidemic dynamics, offering more accurate descriptions than traditional models \cite{krapf2015mechanisms,chen2021review,yu2016fractional}. 
Recent studies combine FDEs with shadow neural networks, primarily applied in computational neuroscience \cite{anastasio1994fractional} and models like Hopfield networks \cite{kaslik2012nonlinear}. These studies primarily engage with numerical simulations and delve into the bifurcation and stability dynamics within such networks.
In recent advancements within deep learning research, \cite{liu2022regularized} have pioneered the application of fractional derivatives for optimizing neural network parameters, offering an alternative to conventional methods such as SGD and Adam \cite{kingma2014adam}. Building on the theoretical underpinnings of fractional calculus, \cite{antil2020fractional} utilize an L1 approximation of fractional derivatives to enhance the architecture of densely connected neural networks, addressing challenges associated with the vanishing gradient problem. 
Additionally, studies on FDE-based Graph Neural Networks (GNNs) \cite{KanZhaDin:C24frond} have employed fractional diffusion and oscillator mechanisms to enhance graph representation learning, achieving superior performance over traditional integer-order models. This approach demonstrates that neural FDEs with fractional derivatives can effectively model the updating and propagation of hidden features over a graph, offering both theoretical and practical advantages.

\section{Preliminaries}\label{sec.prelim}
Fractional calculus excels in modeling systems characterized by non-local interactions, where the system's future state is determined by its extensive historical context.
Here, we provide a succinct introduction to the fundamental concepts of fractional calculus. Throughout this paper, we adhere to standard assumptions that ensure the well-posedness of our formulations, e.g., the validity of integrals and the existence and uniqueness of solutions to differential equations, as detailed in foundational texts \cite{diethelm2009numerical, diethelm2002analysis}.

\subsection{Fractional Calculus}
\noindent\tb{Traditional Calculus:} In traditional calculus, the first-order derivative of a scalar function $z(t)$ quantifies the local rate of change, defined as:
\begin{align}
\frac{\ud z(t)}{\ud t} \equiv \dot{z}(t) \coloneqq \lim _{\Delta t \rightarrow 0} \frac{z(t+\Delta t)-z(t)}{\Delta t}.
\end{align}
We denote by $J$ the operator that maps a function $z$, assumed to be (Riemann) integrable over the compact interval $[0, T]$, to its primitive centered at $0$, i.e.,
\begin{align}
J z(t) \coloneqq \int_0^t z(u) \ud u \quad \text { for } 0 \leq t \leq T.
\end{align}
For any integer $n \in \mathbb{N}^{+}$, the notation $J^n$ represents the $n$-fold iteration of $J$, defined such that $J^1 := J$ and $J^n := J J^{n-1}$ for $n \geq 2$. Equivalently, by using integration by parts, we have \cite{diethelm2010analysis}[Lemma 1.1.]:
\begin{align}
J^n z(t)=\frac{1}{(n-1)!} \int_0^t(t-u)^{n-1} z(u) \ud u \text{ with $n\in \mathbb{N}^{+}$}. \label{eq.fdasf}
\end{align}

\noindent\textbf{Fractional Integrals:} The concept of a fractional integral generalizes the classical integral operator. Two commonly used definitions are the left- and right-sided Riemann-Liouville fractional integrals \cite{tarasov2011fractional}[page 4], denoted by $J_{\mathrm{left}}^\beta$ and $J_{\mathrm{right}}^\beta$ respectively, with $\beta \in \Real^+$. These operators are defined as:
\begin{align}
\begin{aligned}
J_{\mathrm{left}}^\beta z(t) &\coloneqq \frac{1}{\Gamma(\beta)} \int_0^t (t-u)^{\beta-1} z(u) \ud u, \\ \label{eq.dafd}
J_{\mathrm{right}}^\beta z(t) &\coloneqq \frac{1}{\Gamma(\beta)} \int_t^T (u-t)^{\beta-1} z(u) \ud u,
\end{aligned}
\end{align}
where $\Gamma(\beta)$ is the gamma function, which extends the factorial function to real number arguments.
Unlike the integer-order $n$ in traditional integrals \cref{eq.fdasf}, the order $\beta$ in \cref{eq.dafd} can take any positive real value.

\noindent\textbf{Fractional Derivatives:} The fractional derivative extends the concept of differentiation to non-integer orders. The left- and right-sided Riemann-Liouville fractional derivatives ${}_{\mathrm{left}}D_{\mathrm{RL}}^\beta$ and ${}_{\mathrm{right}}D_{\mathrm{RL}}^\beta$ are formally defined as \cite{tarasov2011fractional}[page 385]:
\begin{align}
    \begin{aligned}\label{eq.RL}
   {}_{\mathrm{left}}D_{\mathrm{RL}}^\beta z(t) &\coloneqq \frac{\ud^m}{\ud t^m} J_{\mathrm{left}}^{m-\beta} z(t)\\
   &= \frac{1}{\Gamma(m-\beta)} \frac{\ud^m}{\ud t^m} \int_0^t \frac{z(\tau) d \tau}{(t-\tau)^{\beta-m+1}}, \\ 
    {}_{\mathrm{right}}D_{\mathrm{RL}}^\beta z(t)&\coloneqq(-1)^m \frac{\ud^m}{\ud t^m} J_{\mathrm{right}}^{m-\beta} z(t) \\
    &\frac{(-1)^m}{\Gamma(m-\beta)} \frac{\ud^m}{\ud t^m} \int_t^T \frac{z(\tau) d \tau}{(\tau-t)^{\beta-m+1}},
    \end{aligned}
\end{align}
where $m$ is an integer such that $m-1< \beta \leq m$. 
Similarly, the expressions ${}_{\mathrm{left}}D_{C}^\beta$ and  ${}_{\mathrm{right}}D_{C}^\beta$ represent the left- and right-sided Caputo fractional derivatives, respectively, as detailed in \cite{tarasov2011fractional}[page 386]. They are defined as follows:
\begin{align}
    \begin{aligned}\label{eq.Cap}
    {}_{\mathrm{left}}D_{C}^\beta z(t) & \coloneqq J_{\mathrm{left}}^{m-\beta} \frac{\ud^m}{\ud t^m} z(t),\\
    &=\frac{1}{\Gamma(m-\beta)} \int_0^t \frac{\frac{\ud^m}{\ud \tau^m}z(\tau) \ud \tau }{(t-\tau)^{\beta-m+1}},\\
    {}_{\mathrm{right}}D_{C}^\beta z(t)& \coloneqq  (-1)^m J_{\mathrm{right}}^{m-\beta}  \frac{\ud^m}{\ud t^m} z(t)\\ 
    &=\frac{(-1)^m}{\Gamma(m-\beta)} \int_t^T \frac{\frac{\ud^m}{\ud \tau^m} z(\tau) \ud \tau }{(\tau-t)^{\beta-m+1}}.
    \end{aligned}
\end{align}
From the definitions, it becomes clear that fractional derivatives integrate the historical states of the function through the integral term, emphasizing their non-local, memory-dependent characteristics. Unlike integer-order derivatives that solely represent the local rate of change of the function at a specific point, fractional derivatives encapsulate a broader spectrum of the function's past behavior, providing a richer analysis tool in dynamical systems where history plays a crucial role.
As the fractional order $\beta$ approaches an integer, these fractional operators naturally converge to their classical counterparts \cite{diethelm2010analysis}, ensuring a smooth transition from fractional to traditional calculus. 
For vector-valued functions, fractional derivatives and integrals are defined component-wise across each dimension, similar to the treatment in integer-order calculus.

\subsection{First-Order Neural ODEs}\label{subsec.odeeq}
In a neural ODE layer, the transformation from the initial feature $\bz(0)=\bz_0\in\mathbb{R}^d$ to the output $\bz(T)\in\mathbb{R}^d$ is governed by the differential equation:
\begin{align}
\ddfrac{\bz(t)}{t}=f(t,\bz(t);{\btheta}),
\end{align}
where the function $f$, mapping from $[0, \infty) \times \mathbb{R}^d$ to $\mathbb{R}^d$ with $d$ being the feature dimension, encapsulates the layer's trainable dynamics for updating hidden features, parameterized by $\btheta$.
The trajectory $\bz(t)$ represents the continuous evolution of the system's hidden state.
A notable technical challenge in training neural ODEs involves performing backpropagation. Direct differentiation using automatic differentiation techniques \cite{paszke2017automatic} is feasible but can incur significant memory costs and introduce numerical inaccuracies. To address this issue, \cite{chen2018neural} introduced the adjoint sensitivity method, originally proposed by Pontryagin \cite{pontryagin1962}. This method efficiently computes the gradients of parameters $\btheta$ by constructing an augmented ODE that operates backward in time.

\section{Neural FDE and Adjoint Backpropagation}\label{sec.framework}
In this section, we introduce the neural FDE framework, which utilizes a neural network to parameterize the fractional derivative of the hidden feature state. This approach integrates a continuum of past states into the current state, allowing for rich and flexible modeling of hidden features. We then propose an effective strategy for training the neural FDE by utilizing an augmented FDE in the reverse direction. Additionally, we describe the method for efficiently solving this augmented FDE.

\subsection{Neural FDE}
In our study, we adopt the Caputo fractional derivative in a manner akin to the approach described by \cite{KanZhaDin:C24frond}. 
The dynamics of the hidden units are modeled by the following neural FDE:
\begin{align}
    {}_{\mathrm{left}}D_{C}^\beta \bz(t)=f(t,\bz(t);\btheta), \quad 0<\beta\le 1. \label{eq.FDE}
\end{align}
In this formulation, $f$, parameterized by $\btheta$, is a function that maps $[0, \infty) \times \mathbb{R}^d$ to $\mathbb{R}^d$ and represents the trainable fractional derivatives of the hidden state. The system state $\bz(t)$, starting from the initial condition $\bz(0)=\bz_0$, evolves up to a predetermined terminal time $T$. This terminal state $\bz(T)$ is then used for downstream tasks such as classification or regression. 
For simplicity, we consider only $0<\beta \leq 1$ without loss of generality, as higher orders can be converted to this range \cite{diethelm2010analysis}, akin to how higher integer-order ODEs can be transformed into first-order systems with augmented states.

The computation of $\bz(T)$ is achieved using a forward FDE solver. Classic solvers such as the fractional explicit Adams–Bashforth–Moulton \cite{diethelm2004detailed} and the implicit L1 solver \cite{gao2011compact,sun2006fully} can be employed. These methods demonstrate how time can serve as a continuous analog to discrete layer indices in traditional neural networks, similar to integer-order ODEs \cite{chen2018neural}. To illustrate, we introduce the following iterative method from \cite{diethelm2004detailed} to solve \cref{eq.FDE},  showcasing the dense connection nature of the approach.

\noindent\tb{Predictor:} Let $h$ be a small positive discretization parameter. Consider a uniform grid spanning $[0, T]$ defined by $\set{t_{k} = kh \given k = 0, 1, \dots, N}$, where $h = \frac{T}{N}$. Let $\bz_{k}$ be the numerical approximation of $\bz(t_k)$. The basic predictor approximation is given by:
\begin{align}
\bz_{k}^{\mathrm{P}}= \bz_0+\frac{1}{\Gamma(\beta)} \sum_{j=0}^{k-1} \mu_{j, k} f(t_j,\bz_{j};{\btheta}),\label{eq.pre}
\end{align}
where $\mu_{j, k}= \frac{h^\beta}{\beta}\left((k-j)^\beta-(k-1-j)^\beta\right)$, and $k$ denotes the discrete time index (iteration).

\noindent\tb{Corrector:} The Predictor only provides a rough approximation of the true solution. To refine this approximation, the corrector formula from \cite{diethelm2004detailed}, a fractional variant of the one-step Adams-Moulton method, can be implemented using $\bz_{k}^{\mathrm{P}}$ as follows:
\begin{align}
\begin{aligned}
    \bz_{k}= \bz_0 & +\frac{1}{\Gamma(\beta)} \sum_{j=0}^{k-1} \eta_{j, k} f(t_j,\bz_{j};\btheta) \\
&+\frac{1}{\Gamma(\beta)} \eta_{k, k} f(t_k,\bz_{k}^{\mathrm{P}};\btheta). \label{eq.cor}
\end{aligned}
\end{align}
The coefficients $\eta_{j, k}$ \cite{diethelm2004detailed} are defined as follows: $\eta_{0, k}(\beta) = \frac{h^\beta}{\beta(\beta+1)} \big((k-1)^{\beta+1} - (k-1-\beta)k^\beta\big)$; for $1 \leq j \leq k-1$, $\eta_{j, k}(\beta) = \frac{h^\beta}{\beta(\beta+1)} \big((k-j+1)^{\beta+1} + (k-1-j)^{\beta+1} - 2(k-j)^{\beta+1}\big)$; and $\eta_{k, k}(\beta) = \frac{h^\beta}{\beta(\beta+1)}$.

\subsection{Adjoint Parameter Backpropagation}\label{sec.adjoint}
From the fractional explicit Adams–Bashforth–Moulton formulations \cref{eq.pre,eq.cor}, it is evident that solving the neural FDE during the forward pass involves numerous iterations.
The works \cite{KanZhaDin:C24frond,ZhaKanSon:C24robustfrond} train neural FDEs using the basic automatic differentiation techniques from PyTorch \cite{paszke2019pytorch}. This method is computationally demanding, involving backpropagation through the numerous forward-pass iterations in FDE solvers.
To address this challenge, we derive an adjoint method to compute gradients of the parameters $\btheta$. Echoing the methodology used in integer-order neural ODEs \cite{chen2018neural}, this technique entails solving a secondary, augmented FDE in reverse time. While the complete derivation is detailed in the Appendix due to space constraints, we outline the critical steps here.

Consider a scalar-valued loss function $L(\cdot)$ that depends on the terminal state $\bz(T)$. Our primary goal is to minimize $L(\bz(T))$ with respect to $\btheta$. We aim to compute the gradient $\ddfrac{L\left(\bz(T)\right)}{\btheta}$ for gradient descent.
The strategy is to find a Lagrangian function $\lambda(t)$ to circumvent the direct computation of challenging derivatives such as $\ddfrac{\bz(T)}{\btheta}$ or $\ddfrac{\bz(t)}{\btheta}$. To this end, we consider the following optimization problem:
\begin{align}
\begin{aligned}
\min_{\btheta} \quad & L\left(\bz\left(T\right)\right) \\
\text{subject to} \quad & \begin{aligned}[t]
&F\left( {}_{\mathrm{left}}D_{C}^\beta \bz(t), \bz(t), \btheta, t\right)  \\
&\quad\quad \coloneqq {}_{\mathrm{left}}D_{C}^\beta \bz(t)- f(t, \bz(t);\btheta) = 0 \\
&\text{for all } t \in [0, T],
\end{aligned} \\
& \bz\left(0\right) = \bz_{0}.
\end{aligned}
\end{align}
We define the augmented objective function $\psi$ as 
\begin{align}
\psi=L\left(\bz\left(T\right)\right)-\int_{0}^{T} \lambda(t) F\left( {}_{\mathrm{left}}D_{C}^\beta \bz(t), \bz(t), \btheta, t\right)  \ud t. \label{eq.fdasfd}
\end{align}
Since $F\left( {}_{\mathrm{left}}D_{C}^\beta \bz(t), \bz(t), \btheta, t\right)=0$ for all $t \in [0, T]$, the derivative $\ddfrac{L\left(\bz(T)\right)}{\btheta}$ is the same as $\ddfrac{\psi}{\btheta}$. For the second term on the right-hand side of \cref{eq.fdasfd}, we can get that
\begin{align*}
\begin{aligned}
\int_{0}^{T} \lambda(t) F \ud t 
& =\lambda(T) J_{\mathrm{left}}^{1-\beta} \bz(T) 
-  \bz(0) J_{\mathrm{right}}^{1-\beta} \lambda(0) \\
& \hspace{-0.9cm}+ \int_0^T {}_{\mathrm{right}}D_{C}^\beta \lambda(t) \bz(t) \ud t
- \int_{0}^{T} \lambda f(t, \bz(t);\btheta) \ud t.
\end{aligned}
\end{align*}
The detailed intermediate steps to achieve this are provided in the Appendix due to space constraints. Taking the derivative with respect to $\btheta$, we have
\begin{align}
\begin{aligned}
& \frac{\mathrm{d}}{\mathrm{d} \btheta}\left[\int_{0}^{T} \lambda F \ud t\right] 
=\lambda\left(T\right) \frac{\mathrm{d} J_{\mathrm{left}}^{1-\beta} \bz(T) }{\mathrm{d} \btheta} \\
&\quad  +\int_{0}^{T}\left({}_{\mathrm{right}}D_{C}^\beta \lambda(t) -\lambda(t) \frac{\partial f}{\partial \bz}\right) \ddfrac{\bz}{\btheta} \ud t  
-\int_{0}^{T} \lambda \frac{\partial f}{\partial \btheta} \ud t,
\end{aligned}
\end{align}
where we have used $\frac{\mathrm{d} \bz(0)}{\mathrm{~d} \btheta}=0$ since $\bz_{0}$ is the initial input hidden feature and it is not dependent on $\btheta$.
It follows that 
\begin{align}
&\frac{\mathrm{d} L}{\mathrm{~d} \btheta} = \ddfrac{\psi}{\btheta} 
=\frac{\ud L}{\ud J_{\mathrm{left}}^{1-\beta} \bz(T)} \frac{\mathrm{d} J_{\mathrm{left}}^{1-\beta} \bz(T)}{\mathrm{d} \btheta}
-\frac{\mathrm{d}}{\mathrm{d} \btheta}\left[\int_{0}^{T} \lambda F \ud t\right]\nn
&=\left(\frac{\ud L}{\ud J_{\mathrm{left}}^{1-\beta} \bz(T)} 
- \lambda\left(T\right) \right) \frac{\mathrm{d} J_{\mathrm{left}}^{1-\beta} \bz(T) }{\mathrm{d} \btheta} \nn
& -\int_{0}^{T}\left({}_{\mathrm{right}}D_{C}^\beta \lambda(t) -\lambda(t) \frac{\partial f}{\partial \bz}\right) \ddfrac{\bz}{\btheta} \ud t  
+\int_{0}^{T} \lambda \frac{\partial f}{\partial \btheta} \ud t. \label{eq.fdasfe}
\end{align}

To avoid the direct computation of challenging derivatives such as $\ddfrac{\bz(T)}{\btheta}$ or $\ddfrac{\bz(t)}{\btheta}$, we let $\lambda(t)$ satisfy the following FDE:
\begin{align}
\begin{aligned}
    {}_{\mathrm{right}}D_{C}^\beta \lambda(t) & = \lambda(t) \frac{\partial f}{\partial \bz}, \\
    \text { with } \lambda\left(T\right) &=\frac{\ud L}{\ud J_{\mathrm{left}}^{1-\beta} \bz(T)}.
\end{aligned} \label{eq.back_adjoint}
\end{align}
Consequently, as the first two terms in \cref{eq.fdasfe} vanish, we obtain
\begin{align}
    \frac{\mathrm{d} L}{\mathrm{~d} \btheta}=-\int_{T}^{0} \lambda(t) \frac{\partial f}{\partial \btheta} \ud t. \label{eq.back_para}
\end{align}
To facilitate computation, we approximate the constraint on the last time point as
$\frac{\ud L}{\ud J_{\mathrm{left}}^{1-\beta} \bz(T)} \approx \frac{\ud L}{\ud \bz(T)}$ in \cref{eq.back_adjoint}. 
This approximation represents the gradient of the loss with respect to the final state of the system. 
Efficient evaluation of the vector-Jacobian products, $\lambda(t) \frac{\partial f}{\partial \bz}$ and $\lambda(t) \frac{\partial f}{\partial \btheta}$, specified in \cref{eq.back_adjoint} and \cref{eq.back_para}, is achieved using automatic differentiation, offering a computational cost on par with that of evaluating $f$ directly \cite{griewank2003mathematical}.

In the reverse model, the systems described in \cref{eq.back_adjoint,eq.back_para} are computed simultaneously. Echoing the methodologies employed in \cref{eq.pre,eq.cor}, we utilize standard quadrature techniques to solve these equations. Our numerical iterations confirm that the coefficients are consistent with those reported in \cref{eq.pre}. Additionally, the trajectory $\bz(t)$ generated during the forward pass can be efficiently reused. The complete numerical scheme is detailed in \cref{ssec.reverse_fde}. \cref{alg1} outlines the procedure for constructing the required dynamics and applying the solver described in \cref{ssec.reverse_fde} to simultaneously compute all gradients.

\begin{algorithm}
\caption{Reverse-mode Differentiation for a Neural FDE}
\label{alg1}
\begin{algorithmic}[1]
\State \textbf{Input:} Dynamics parameters $\btheta$, initial time $0$, final time $T$, trajectory $\bz(t)$, loss gradient $\ud L / \ud \bz(T)$

\State \textbf{Objective:} Compute the gradients  $\ddfrac{L}{\btheta}$ using reverse-time integration.

\State \textbf{Initialize:} Set $\lambda(T) = \frac{\ud L}{\ud \bz(T)}$ and $\frac{\ud L}{\ud \btheta}(T) = \mathbf{0}_{|\btheta|}$.

\State Solve the reverse-time FDE for $\lambda(t)$:
\begin{align*}
    {}_{\mathrm{right}}D_{C}^\beta \lambda(t) &= \lambda(t) \frac{\partial f}{\partial \bz}, \quad \text{(Adjoint equation)}
\end{align*}

\State Simultaneously compute the gradient with respect to parameters $\btheta$:
\begin{align*}
    \frac{{\mathrm{d} L}/{\mathrm{~d} \btheta}}{\ud t} &= \lambda(t) \frac{\partial f}{\partial \btheta}, \quad \text{(Parameter sensitivity)}
\end{align*}

\State \textbf{Output:} Return the computed gradient $\ddfrac{L}{\btheta}$ upon completion of the integration from $T$ to $0$.

\end{algorithmic}
\end{algorithm}

\subsection{Solving the Reverse-Mode FDE} \label{ssec.reverse_fde}
To solve the equations described in \cref{eq.back_adjoint,eq.back_para}, we first convert \cref{eq.back_adjoint} into the corresponding Volterra integral equation:
\begin{align}
    \lambda(t) = \lambda(T) + \frac{1}{\Gamma(\beta)} \int_t^T (s-t)^{\beta-1} \lambda(s) \frac{\partial f}{\partial \bz} \ud s.
\end{align}
Consider a small positive discretization parameter $h$, and a uniform grid over the interval $[0, T]$, defined by $\{t_k = kh \mid k = 0, 1, \dots, N\}$, where $h = \frac{T}{N}$. Let $\lambda_k$ denote the numerical approximation of $\lambda(t_k)$.
Using the product rectangle rule and initializing with $\lambda_N = \lambda(T)$, we derive the following basic predictor to iteratively compute $\lambda_k$:
\begin{align}
    \lambda_{N-k-1} = \lambda(T) + \frac{1}{\Gamma(\beta)} \sum_{j=N-k}^N b_{j, k+1} \lambda_j \frac{\partial f}{\partial \bz_j}, \label{eq.back_pred1}
\end{align}
where the coefficients $b_{j,k+1}$ are given by:
\begin{align*}
    b_{j,k+1} = \frac{h^\beta}{\beta} \left((j-(N-k-1))^\beta - (j-(N-k))^\beta\right).
\end{align*}
Note that the full discretized trajectory $\{\bz_j\}$, obtained from the forward pass in \cref{eq.pre,eq.cor}, can be reused during the backward computation.

\begin{figure}[H]
    \centering
    \includegraphics[width=0.9\linewidth]{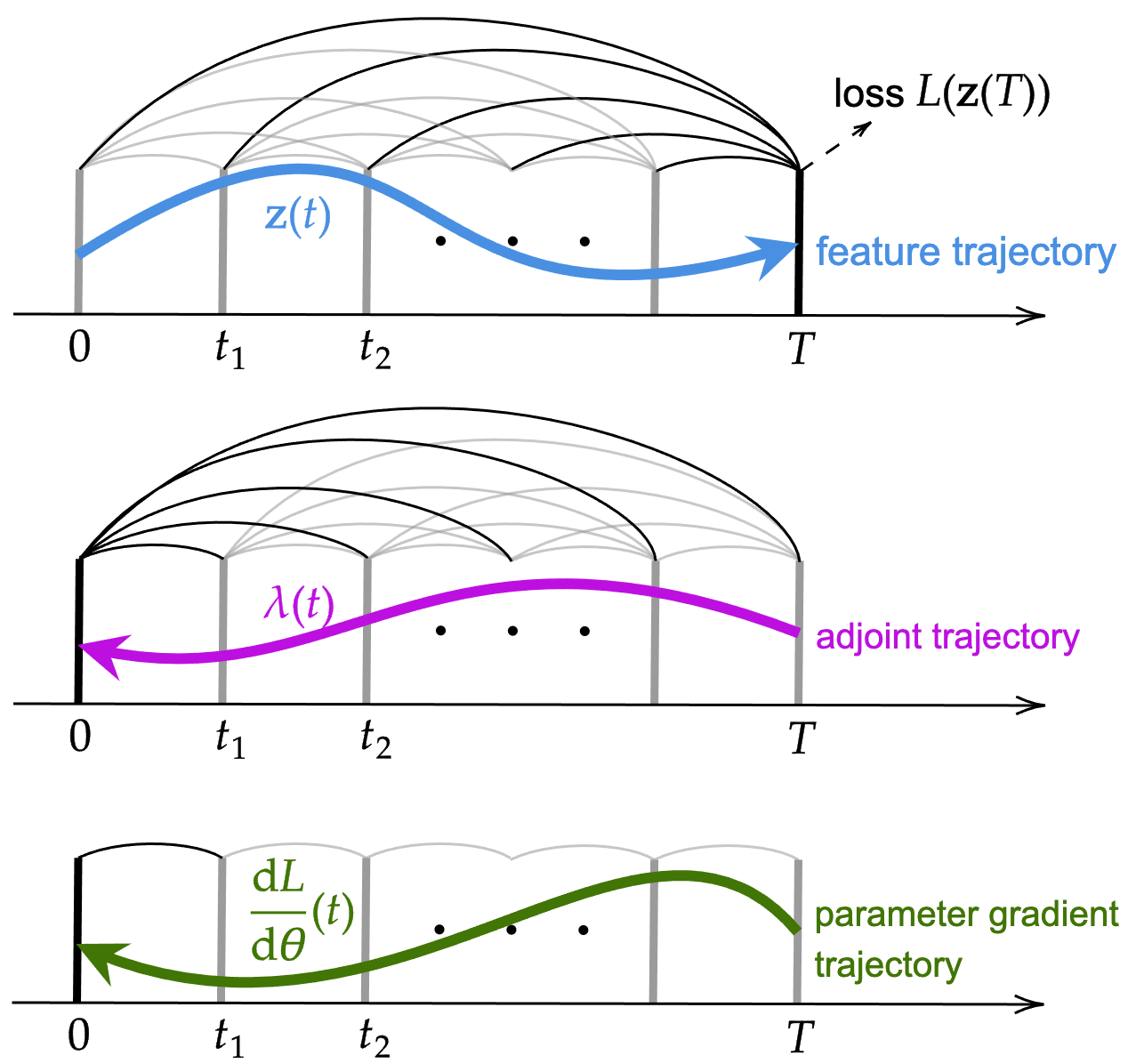}
    \caption{The visualization of the adjoint backpropagation method for training neural FDEs by solving an augmented FDE backward in time.
    }
    \label{fig:adjoint}
\end{figure}

For the integration in \cref{eq.back_para}, we employ a basic Euler scheme using the same uniform time grid. Let us denote the numerical approximation of $\frac{\ud L}{\ud \btheta}(t_k)$ as $g^{\btheta}_k$, initialized as $g^{\btheta}_{N} = \mathbf{0}_{|\btheta|}$, where $\mathbf{0}_{|\btheta|}$ denotes a vector of zeros with the same dimensionality as $\btheta$. We then compute:
\begin{align}
    g^{\btheta}_{N-k-1} = g^{\btheta}_{N-k} + \lambda_{N-k} \frac{\partial f}{\partial \btheta}.\label{eq.back_pred2}
\end{align}
Finally, we obtain $\frac{\mathrm{d} L}{\mathrm{d} \btheta}\approx g^{\btheta}_{0}$ and will be used as the gradient for backpropagation. While advanced corrector formulas could potentially offer more accurate integration to compute the gradient, this work only considers the basic predictor to solve the reverse-mode FDE \cref{eq.back_adjoint,eq.back_para}.
The adjoint backpropagation method is illustrated in \cref{fig:adjoint}, offering a visual depiction of the gradient computation process within the reverse-mode framework.

\begin{table*}[!ht]
 \centering
\fontsize{9pt}{10pt}\selectfont
\setlength{\tabcolsep}{1pt}
\makebox[0.5\textwidth][c]{
 \begin{tabular}{l|ccccccccc}
\toprule
Method   & Cora      & Citeseer           &  Pubmed      & CoauthorCS             & Computer            & Photo          & CoauthorPhy  & Ogbn-arxiv   \\

\midrule
GCN \cite{kipf2017semi}    &  81.5$\pm$1.3        &  71.9$\pm$1.9   &  77.8$\pm$2.9   &  91.1$\pm$0.5     & 82.6$\pm$2.4  &  91.2$\pm$1.2   &  92.8$\pm$1.0  &    72.2$\pm$0.3    \\

GAT \cite{velickovic2018gat}  &  81.8$\pm$1.3       &  71.4$\pm$1.9   &  78.7$\pm$2.3   &  90.5$\pm$0.6  &  78.0$\pm$19.0  & 85.7$\pm$20.3   &  92.5$\pm$0.9  &  \first{73.7$\pm$0.1}   \\
HGCN \cite{chami2019hyperbolic_GCNN}&  78.7$\pm$1.0 &  65.8$\pm$2.0 &  76.4$\pm$0.8  & 90.6$\pm$0.3   & 80.6$\pm$1.8   & 88.2$\pm$1.4   & 90.8$\pm$1.5 & OOM  \\

CGNN \cite{xhonneux2020continuous}&  81.4$\pm$1.6 &  66.9$\pm$1.8 &  66.6$\pm$4.4  & 92.3$\pm$0.2   & 80.3$\pm$2.0   & 91.4$\pm$1.5   & 91.5$\pm$1.8 &  58.7$\pm$2.5 \\

GDE \cite{poli2019graph}&  78.7$\pm$2.2 &  71.8$\pm$1.1 &  73.9$\pm$3.7  & 91.6$\pm$0.1   &  82.9$\pm$0.6   & 92.4$\pm$2.0   & 91.3$\pm$1.1 &  56.7$\pm$10.9 \\

\midrule
GRAND-l    &  {83.6$\pm$1.0 }       &  73.4$\pm$0.5   &  78.8$\pm$1.7   & 92.9$\pm$0.4     & 83.7$\pm$1.2  & 92.3$\pm$0.9   &  93.5$\pm$0.9  &  71.9$\pm$0.2   \\

GRAND-nl    &  82.3$\pm$1.6        &  70.9$\pm$1.0   &  77.5$\pm$1.8   & 92.4$\pm$0.3     &  82.4$\pm$2.1  & 92.4$\pm$0.8   &  91.4$\pm$1.3    &  71.2$\pm$0.2  \\
F-GRAND-l   & \second{84.8$\pm$1.1}    & {74.0$\pm$1.5}  & \second{79.4$\pm$1.5}  & \second{93.0$\pm$0.3}  & {84.4$\pm$1.5}  & \second{92.8$\pm$0.6}  & \first{94.5$\pm$0.4} &  72.6$\pm$0.1  \\

F-GRAND-nl   &  83.2$\pm$1.1    & \second{74.7$\pm$1.9}  & {79.2$\pm$0.7}  & {92.9$\pm$0.4}  & {84.1$\pm$0.9}  & {93.1$\pm$0.9}  & {93.9$\pm$0.5} & 71.4$\pm$0.3  \\

adj-F-GRAND-l   &  \first{85.0$\pm$1.0}    &  \first{75.0$\pm$1.3} &  \first{79.7$\pm$1.6} & \first{93.1$\pm$0.3}  & \second{86.9$\pm$1.4}  &  \first{93.3$\pm$0.5}  & \second{94.0$\pm$0.5}   &  72.5$\pm$0.3  \\

adj-F-GRAND-nl   &   82.6$\pm$1.3   & 74.6$\pm$1.9 & 78.5$\pm$1.5   & 92.8$\pm$0.3  & \first{87.5$\pm$0.8} & 92.5$\pm$0.8  & 93.8$\pm$0.6   &  \second{72.6$\pm$0.3}  \\

\bottomrule
\end{tabular}}
\caption{Node classification results(\%) for random train-val-test splits. The best and the second-best results for each criterion are highlighted in bold and underlined, respectively.}
\label{tab:noderesults}
\end{table*}

\subsection{Model Complexity}\label{ssec.complex}
During the forward pass, as described in \cref{eq.pre}, the process requires computing the fractional derivative $f(t_k,\bz_k;\btheta)$ at each iteration, with results stored in memory for subsequent steps.
The total time complexity over the entire process can be expressed as $\sum_{k=0}^N (C + O(k))$, where $O(k)$ represents the computational overhead of summing and weighting the $k$ terms at each step. Here, $N = \frac{T}{h}$ denotes the number of discretization (iteration) steps necessary for the integration process, and $C$ indicates the computational complexity of the function $f$. 
This leads to a total cost of $O(NC + N^2)$. 
With a fast algorithm for the convolution computations, we generally need $O(N\log(N))$ for the convolution \cite{mathieu2013fast}, resulting in $O(NC + N\log(N))$. 
The memory complexity is represented by $O(P+Nd)$, where $d$ indicates the memory requirement for each hidden state $\bz_k$ and $P$ denotes the peak memory usage for computing $f$ at a single timestep.
The term $Nd$ arises from storing all fractional derivatives $\{f(t_j,\bz_{j};{\btheta})\}_{j=1}^N$ as described in \cref{eq.pre}, where each value has the same dimension as the state $\bz_k$.

During the backward phase, as indicated by computations in \cref{eq.back_pred1,eq.back_pred2}, a sequence of vector-Jacobian products is required. According to \cite{griewank2003mathematical}, computing these products demands approximately at most 2 to 3 times the computational time complexity compared to evaluating the original function $f$. The time complexity is again $O(NC + N^2)$. 
The computational memory complexity stands at $O(Q+Nd)$, assuming the memory cost for the vector-Jacobian product $\lambda \frac{\partial f}{\partial \btheta}$ is around $Q$. The term $d$  arises because $\lambda \frac{\partial f}{\partial \bz}$ shares the same dimension as the hidden state.
For large-dimensional $\btheta$, the gradient memory requirement for $\lambda \frac{\partial f}{\partial \btheta}$ may be the dominant one since the dimension of $\btheta$ typically exceeds that of the hidden state $\bz_k$ in practical implementations.

Direct differentiation through the forward iterations \cref{eq.pre} requires storing both intermediate states and their gradients with respect to all model parameters at each time step since the computation involves all values from past time steps with a dense connection pattern. 
This is essential for calculating gradients by applying the chain rule throughout the intermediate computations.
The memory requirement increases to at least $O(NQ+Nd)$ assuming the gradient memory requirement at each time step is around $Q$, thereby exceeding $O(Q+N d)$ required by our adjoint method. 

A final note is that the memory requirement $O(Nd)$ for storing all fractional derivatives $\{f(t_j,\bz_{j};{\btheta})\}_{j=1}^N$ can be reduced to $KO(d)$ by using the short memory principle  \cite{deng2007short,podlubny1999fractional} to modify the summation in \cref{eq.back_pred1} to $\sum^{N-k+K-1}_{j=N-k}$.  This approximation corresponds to using a shifting memory window with a fixed width $K$ rather than the full history. The diagram is shown in \cite{KanZhaDin:C24frond}[Figure 1].

\begin{table*}[!htp]
 \centering
\fontsize{9pt}{10pt}\selectfont
\setlength{\tabcolsep}{1pt}
\makebox[\textwidth][c]{
    \begin{tabular}{c|ccccccccc}
    \toprule
        Model &  MLP & Node2vec & GCN & GraphSAGE & GRAND-l & F-GRAND-l & F-GRAND-nl  & adj-F-GRAND-l   & adj-F-GRAND-nl \\
        \midrule
        Accuracy &   61.06$\pm$0.08 & 72.49$\pm$0.10 & 75.64$\pm$0.21 & 78.29$\pm$0.16 & 75.56$\pm$0.67 & 77.25$\pm$0.62 & 77.01$\pm$0.22 & \first{78.36$\pm$0.32}  & \second{78.33$\pm$0.20}  \\
    \bottomrule     
    \end{tabular}}
     \caption{Node classification accuracy(\%) on Ogbn-products dataset. The best and the second-best results for each criterion are highlighted in bold and underlined, respectively.}
    \label{tab:ogbn-products}
\end{table*}

\section{Experiments} \label{sec.exp}
To evaluate the efficiency of our neural FDE solvers, experiments were conducted on three tasks: biological system FDE discovery, image classification, and graph node classification. The experiments in this section are designed to achieve two main objectives: 1) To verify that our adjoint FDE training accurately computes gradients and supports backpropagation. This is demonstrated in the small-scale problem described in \cref{ssec.lotka}, where the estimated parameters are shown to converge to the ground-truth values following the adjoint gradients. 2) To show that our adjoint FDE training is memory-efficient for large-scale problems. Experiments in \cref{ssec.image,ssec.exp_graph}, particularly with the large-scale Ogbn-Products dataset, support this claim. It is important to note that our primary goal is to showcase the efficiency of the proposed adjoint backpropagation method for training neural FDEs by solving the augmented FDE backward in time, rather than achieving state-of-the-art results. Our empirical tests demonstrate that this approximation not only reduces the computational memory required for training but also maintains reasonable performance across various experimental setups. 
The experiments were conducted on a workstation running Ubuntu 20.04.1, equipped with an AMD Ryzen Threadripper PRO 3975WX with 32 cores and an NVIDIA RTX A5000 GPU with 24GB of memory.

\subsection{Fractional Lotka-Volterra Model}\label{ssec.lotka}
We consider a nonlinear fractional Lotka-Volterra system, comprising two differential equations that describe the dynamics of biological systems where two species interact, one as a predator and the other as a prey:
\begin{align*}
\begin{aligned}
& {}_{\mathrm{left}}D_{C}^\beta x=x(a-c y), \\
& {}_{\mathrm{left}}D_{C}^\beta y=-y(b-d x),
\end{aligned}
\end{align*}
where $x$ and $y$ represent prey and predator populations, respectively, and $a, b, c, d$ are constants indicating interaction dynamics. We set the ground truth parameters as $[a,b,c,d]=[1.0,0.5,1.0,0.3]$. Using synthetic data generated with these parameters and initial conditions randomly selected from $[0.5,5]$, we train a model to estimate the parameters. The model uses the Adam optimizer \cite{kingma2014adam} with a learning rate of 0.01. After 30 epochs, the estimated parameters, $[0.99, 0.48, 1.05, 0.33]$, closely match the true values, demonstrating the efficiency of our adjoint backpropagation.
We do not validate the gradient in high-dimensional cases because these scenarios have many local minima, and there is no guarantee of achieving the global minimum.

\begin{table}[!t]
 \centering
\fontsize{9pt}{10pt}\selectfont
\setlength{\tabcolsep}{2pt}
\makebox[0.4\textwidth][c]{
    \begin{tabular}{c|c|c|c|c|c}
    \toprule
       Method  & Test Error &  \makecell{Train. GPU \\ Mem (MB)} & \makecell{Training \\ Time (s)} & \makecell{Inf. GPU \\Mem (MB)} & \makecell{Inference \\Time (s)} \\
    \midrule
    Direct &  0.39\%  & 3612 &1.46 & 1628 &0.54\\
   Adjoint &  0.36\% &  2432 & 1.41& 1628 & 0.54\\
    \bottomrule
    \end{tabular}
    }
     \caption{Comparative performance of direct differentiation vs. adjoint backpropagation on the MNIST dataset with  $T=20$.}
    \label{tab:mnist}
\end{table}

\begin{table}[!t]
\centering
\fontsize{9pt}{10pt}\selectfont
\setlength{\tabcolsep}{1.8pt}
\makebox[0.4\textwidth][c]{
\begin{tabular}{c|c|c|c|c|c|c|c|c|c|c|c}
    \toprule
$T$ & 1 & 5 & 10 & 20 & 30 & 40 & 50 & 60 & 70 & 100 & 200 \\
\hline
$\text{M}_{\mathrm{dir}}$ & 1.8 & 2.2 & 2.7 & 3.6 & 4.6 & 5.5 & 6.5 & 7.4 & 8.4 & 11.2 & 20.7  \\

$\text{M}_{\mathrm{adj}}$ & 1.8 & 1.9 & 2.1 & 2.4 & 2.5 & 2.9 & 3.2 & 3.5 & 3.9 & 5.0 & 8.4 \\
$\frac{\text{M}_{\mathrm{adj}}}{\text{M}_{\mathrm{dir}}}$ & 1.00 & 0.88 & 0.79 & 0.67 & 0.55 & 0.51 & 0.49 & 0.48 & 0.46 & 0.44 & 0.40 \\
    \bottomrule
    \end{tabular}
    }
    \caption{GPU memory usage (GB) during training on the MNIST dataset across different integral times $T$. Notations $\text{M}_{\mathrm{dir}}$ and $\text{M}_{\mathrm{adj}}$ represent the memory usage using the direct differentiation and the adjoint backpropagation. The ration $\frac{\text{M}_{\mathrm{adj}}}{\text{M}_{\mathrm{dir}}}$ quantifies the relative memory consumption between the two methods.}
    \label{tab:mnist_T}
\end{table}

\subsection{Image Classification}\label{ssec.image}
This experiment evaluates the performance of neural FDEs on the MNIST dataset \cite{lecun1998gradient}, with a focus on comparing the adjoint backpropagation to the direct differentiation through forward-pass iterations. The aim is to assess the models in terms of both accuracy and computational cost. Following the model architecture from \cite{chen2018neural}, the input is downsampled twice.
The model's hidden features are then updated following a neural FDE. Here, $f(t,\bz(t);\btheta)$ is configured as a convolution module. 
The step size $h$ is set to $0.1$, and the fractional order $\beta$ is set to $0.5$. 
This configuration implies a continuous analog to discrete layer indices in traditional neural networks, corresponding to approximately $N=T/h$ layers.
Subsequently, a fully connected layer is applied to the extracted features for classification.  In both the training and testing phases, the batch size is set to 128.
In the first setting, gradients are backpropagated directly through forward-pass operations in \cref{eq.pre}, while in the second setting, we solve the proposed reverse-mode FDE in \cref{ssec.reverse_fde} to obtain the gradients.

The model's test accuracy, along with training and testing memory and time, are presented in \cref{tab:mnist} when $T=20$. Both training methods achieve an accuracy of over 99.5\%. Comparing the training memory, we find that training neural FDEs by solving the augmented FDE backward reduces memory usage by nearly 33\% in this setting.
Furthermore, we set integral times $T$ ranging from 1 to 200 and record the GPU memory usage in \cref{tab:mnist_T}. We observe that with sufficiently large $T$, the adjoint backpropagation consumes only 40\% of the training memory compared to the direct differentiation.

\begin{table}[!b]
  \centering
\fontsize{9pt}{10pt}\selectfont
\setlength{\tabcolsep}{1pt}
\makebox[0.45\textwidth][c]{
    \begin{tabular}{c|c|c|c|c}
    \toprule
       Model  & \makecell{adj-F-\\GRAND-l} & F-GRAND-l  & \makecell{adj-F-\\GRAND-nl} & F-GRAND-nl \\
    \midrule
       \makecell{Inference \\Time (s)}   &  0.102 & 0.102  & 0.185 & 0.185 \\
       \midrule
       \makecell{Inf. GPU \\ Mem. (MB)} & 3982 & 3982  & 4314 & 4314 \\
       \midrule
        \makecell{Training \\Time (s)}   & 0.319  & 0.352  &  0.785 &  0.806\\
        \midrule
       \makecell{Train. GPU \\Mem. (MB)}  & 5570 &  8527 & 9086 & 18180 \\
    \bottomrule
    \end{tabular}
    }
       \caption{Computation cost of models on the Ogbn-arxiv dataset: integral time $T=10$ and step size of 1.}
    
    \label{tab:model_arxiv}
\end{table}

\begin{table}[!b]
  \centering
\fontsize{9pt}{10pt}\selectfont
\setlength{\tabcolsep}{1pt}
\makebox[0.45\textwidth][c]{
    \begin{tabular}{c|c|c|c|c}
    \toprule
       Model  & \makecell{adj-F-\\GRAND-l} & F-GRAND-l  & \makecell{adj-F-\\GRAND-nl} & F-GRAND-nl \\
    \midrule
       \makecell{Inference \\Time (s)}  & 34.39 & 34.39  &  36.11&  36.11 \\
       \midrule
       \makecell{Inf. GPU \\ Mem. (MB)} & 2678  & 2678  & 4468 & 4468  \\
        \midrule
        \makecell{Training \\Time (s)}  &  34.04 & 35.15  & 43.24 & 44.41 \\
         \midrule
      \makecell{Train. GPU \\Mem. (MB)}  & 6602   &  7950 & 11238 & 17210 \\
    \bottomrule
    \end{tabular}
    }
    \caption{Computation cost of models on the Ogbn-products dataset: integral time $T=10$ and step size of 1.}
    \label{tab:model_products}
\end{table}
\subsection{Node Classification on Graph Dataset}\label{ssec.exp_graph}
In this section, we validate the efficiency of our neural FDE solvers through experiments conducted across a range of graph node classification tasks, as outlined in \cite{KanZhaDin:C24frond}. 
These experiments utilize the neural FDE model F-GRAND, detailed in \cite{KanZhaDin:C24frond}[Sec. 3.1.1].
It includes two variants, F-GRAND-nl and F-GRAND-l, denoting nonlinear and linear graph feature dynamics $f$, respectively. Utilizing our neural FDE toolbox, we aim to demonstrate that our solver can match the performance of traditional solvers \cite{KanZhaDin:C24frond} that rely on the direct differentiation using PyTorch without using the adjoint backpropagation method. Models trained using our adjoint backpropagation technique are prefixed with adj-. Moreover, we show that our neural FDE toolbox significantly reduces computational memory demand during training.

We follow the experimental setup from GRAND \cite{chamberlain2021grand}, conducting experiments on homophilic datasets.
We adopt the same dataset splitting method as in \cite{chamberlain2021grand}, using the Largest Connected Component (LCC) and performing random splits. For the Ogbn-products dataset, we employ a mini-batch training approach as outlined in the paper \cite{zeng2020graphsaint}. For detailed information on the dataset and implementation specifics, please refer to the Appendix.

From \cref{tab:noderesults}, we observe that our adjoint backpropagation delivers comparable performance across all datasets on node classification tasks. This demonstrates the effectiveness of our proposed gradient computation using the proposed reverse-mode FDE.
Furthermore, \cref{tab:ogbn-products} includes the large-scale Ogbn-Products dataset, with 2449029 nodes and 61859140 edges. The memory efficiency of adj-F-GRAND enables the use of larger batch sizes, which contributes to improved classification outcomes. 
In our experiments, the batch size for adj-F-GRAND is set to 20,000 compared to 10,000 for F-GRAND when executed on the same GPU. Setting the F-GRAND batch size to 20000, however, leads to Out-Of-Memory (OOM) errors.

We also investigate the computational memory costs associated with these training methods with the other settings all the same.
From \cref{tab:model_arxiv} and \cref{tab:model_products}, it is evident that our adjoint solvers significantly reduce computational memory costs during the training phase for the same model. 
Especially for the GRAND-nl model, which recomputes the attention score at each integration step, our adjoint solvers require only half the memory compared to traditional solvers. 
This highlights the remarkable efficiency of our adjoint method.

\section{Conclusion}\label{sec.conclusion}
In this paper, we propose an efficient neural FDE training strategy by solving an augmented FDE backward in time, which substantially reduces memory requirements. Our approach provides a practical neural FDE toolbox and holds considerable promise for diverse applications. We demonstrate the effectiveness of our solver in image classification, biological system FDE discovery, and graph node classification. Our training using the adjoint backpropagation can perform comparably to baseline models while significantly reducing computational overhead. The new neural FDE training technique will benefit the community by enabling more efficient use of computational resources and has the potential to scale to large FDE systems.

\section*{Acknowledgments}
This research is supported by the National Research Foundation, Singapore and Infocomm Media Development Authority under its Future Communications Research and Development Programme. It is also supported by the National Natural Science Foundation of China under Grant Nos. 12301491, 12225107 and 12071195, the Major Science and Technology Projects in Gansu Province-Leading Talents in Science and Technology under Grant No. 23ZDKA0005, the Innovative Groups of Basic Research in Gansu Province under Grant No. 22JR5RA391, and Lanzhou Talent Work Special Fund. 
To improve the readability, parts of this paper have been grammatically revised using ChatGPT \cite{openai2022chatgpt4}.

\bibliography{IEEEabrv,StringDefinitions,refs}
\newpage

\appendix

This supplementary document complements the main paper by providing comprehensive details and supporting evidence necessary for a full understanding of the research conducted. 
The sections are organized as follows:

\begin{enumerate}
    \item A detailed exposition on the derivation of the adjoint method is provided in \cref{sec.supp_adjoint_full}. This section outlines the theoretical underpinnings and computational efficiencies gained through our approach, enhancing the reader's grasp of our method.
    \item The experimental design and dataset specifics are extensively covered in \cref{sec.supp_exp}. This includes a detailed breakdown of dataset characteristics, preprocessing steps, and experimental setups.
    \item  We discuss the limitations of our work and discuss its broader impact in  \cref{sec.supp_limit}.
\item Lastly, the entire codebase used in our research is made available. This includes the implementation of the adjoint backpropagation technique within the \texttt{torchfde} toolbox, available at \url{https://github.com/kangqiyu/torchfde}. 
\end{enumerate}

\section{Neural FDE and Adjoint Backpropagation}\label{sec.supp_adjoint_full}
In the main paper \cref{sec.adjoint}, we sketch the derivation of the adjoint method which is an effective strategy for training the neural FDE by utilizing an augmented FDE in the reverse direction. 
In this section, we present the full derivation with more discussions.

Consider a scalar-valued loss function $L(\cdot)$ that depends on the terminal state $\bz(T)$. Our primary goal is to minimize $L(\bz(T))$ with respect to $\btheta$. We need to compute the gradient $\ddfrac{L\left(\bz(T)\right)}{\btheta}$ for gradient descent.
The strategy is to find a Lagrangian function $\lambda(t)$ to circumvent the direct computation of challenging derivatives such as $\ddfrac{\bz(T)}{\btheta}$ or $\ddfrac{\bz(t)}{\btheta}$. To this end, we consider the following optimization problem:
\begin{align}
\begin{aligned}
\min_{\btheta} \quad & L\left(\bz\left(T\right)\right) \\
\text{subject to} \quad & \begin{aligned}[t]
&F\left( {}_{\mathrm{left}}D_{C}^\beta \bz(t), \bz(t), \btheta, t\right)  \\
&\quad\quad \coloneqq {}_{\mathrm{left}}D_{C}^\beta \bz(t)- f(t, \bz(t);\btheta) = 0 \\
&\text{for all } t \in [0, T],
\end{aligned} \\
& \bz\left(0\right) = \bz_{0}.
\end{aligned}
\end{align}
We define the augmented objective function $\psi$ as 
\begin{align}
\psi=L\left(\bz\left(T\right)\right)-\int_{0}^{T} \lambda(t) F\left( {}_{\mathrm{left}}D_{C}^\beta \bz(t), \bz(t), \btheta, t\right)  \ud t. \label{eq.supp_fdasfd}
\end{align}
Since $F\left( {}_{\mathrm{left}}D_{C}^\beta \bz(t), \bz(t), \btheta, t\right)=0$ for all $t \in [0, T]$, the derivative $\ddfrac{L\left(\bz(T)\right)}{\btheta}$ is the same as $\ddfrac{\psi}{\btheta}$.

For the second term on the right-hand side of \cref{eq.supp_fdasfd}, we have
\begin{align}
&\int_{0}^{T} \lambda(t) F \ud t \nn
& =\int_{0}^{T} \lambda(t)( {}_{\mathrm{left}}D_{C}^\beta \bz(t)-f(t, \bz(t);\btheta)) \ud t \nn
& =\int_{0}^{T} \lambda(t)  {}_{\mathrm{left}}D_{C}^\beta \bz(t) \ud t-\int_{0}^{T} \lambda(t) f(t, \bz(t);\btheta) \ud t.  \label{eq.supp_feafdfv}
\end{align}
For the first term in \cref{eq.supp_feafdfv}, we have
\begin{align*}
&\int_{0}^{T} \lambda(t)  {}_{\mathrm{left}}D_{C}^\beta \bz(t)) \ud t \nn
& =  \frac{1}{\Gamma(1-\beta)} \int_{0}^{T} \lambda(t)  \int_0^t \dot{\bz}(u)(t-u)^{-\beta} \mathrm{d} u \ud t \\
& =   \frac{1}{\Gamma(1-\beta)} \int_0^T \dot{\bz}(u) \int_u^T \lambda(t)(t-u)^{-\beta} \ud t \ud u \\
& =   \frac{1}{\Gamma(1-\beta)} \left.\bz(u)\left(\int_u^T \lambda(t)(t-u)^{-\beta} d t\right)\right|_0 ^T \\
&\quad \quad -\frac{1}{\Gamma(1-\beta)} \int_0^T \bz(u) \frac{\ud}{\ud u} \left(\int_u^T \lambda(t)(t-u)^{-\beta} \ud t\right) \ud u \\
& =   - \frac{\bz(0)}{\Gamma(1-\beta)} \int_0^T \lambda(t)t^{-\beta} \ud t \\
&\quad \quad -\frac{1}{\Gamma(1-\beta)} \int_0^T \bz(u) \frac{\ud}{\ud u} \left(\int_u^T \lambda(t)(t-u)^{-\beta} \ud t\right) \ud u \\
& =   - \frac{\bz(0)}{\Gamma(1-\beta)} \int_0^T \lambda(t)t^{-\beta} \ud t
+\int_0^T  \bz(u) {}_{\mathrm{right}}D_{\mathrm{RL}}^\beta \lambda(u) \ud u \\
& =   - \frac{\bz(0)}{\Gamma(1-\beta)} \int_0^T \lambda(t)t^{-\beta} \ud t \\
&\quad \quad +\int_0^T \left( {}_{\mathrm{right}}D_{C}^\beta \lambda(u) + \frac{(T-u)^{-\beta}}{\Gamma(1-\beta)} \lambda(T)\right)\bz(u) \ud u \\
& =   - \frac{\bz(0)}{\Gamma(1-\beta)} \int_0^T \lambda(t)t^{-\beta} \ud t 
+\frac{\lambda(T)\int_0^T \bz(u)(T-u)^{-\beta} \ud u}{\Gamma(1-\beta)}  \\
&\quad \quad +\int_0^T {}_{\mathrm{right}}D_{C}^\beta \lambda(u) \cdot  \bz(u) \ud u \\
& = \lambda(T) J_{\mathrm{left}}^{1-\beta} \bz(T) - \bz(0) J_{\mathrm{right}}^{1-\beta} \lambda(0) \\
&\quad\quad +\int_0^T {}_{\mathrm{right}}D_{C}^\beta \lambda(u)\cdot \bz(u) \ud u. 
\end{align*}
Combining \cref{eq.supp_fdasfd} with the above results, we derive:
\begin{align*}
\begin{aligned}
\int_{0}^{T} \lambda(t) F \ud t 
& =\lambda(T) J_{\mathrm{left}}^{1-\beta} \bz(T) 
-  \bz(0) J_{\mathrm{right}}^{1-\beta} \lambda(0) \\
& \hspace{-0.9cm}+ \int_0^T {}_{\mathrm{right}}D_{C}^\beta \lambda(t) \bz(t) \ud t
- \int_{0}^{T} \lambda f(t, \bz(t);\btheta) \ud t.
\end{aligned}
\end{align*}
Take the derivative with respect to $\btheta$, we have
\begin{align*}
    \begin{aligned} 
   \frac{\mathrm{d}}{\mathrm{d} \btheta}\left[\int_{0}^{T} \lambda F \ud t\right]
    &  =\lambda\left(T\right) \frac{\mathrm{d} J_{\mathrm{left}}^{1-\beta} \bz(T) }{\mathrm{d} \btheta}
    -J_{\mathrm{right}}^{1-\beta} \lambda(0) \frac{\mathrm{d} \bz(0)}{\mathrm{d} \btheta}\\
  &   +\int_{0}^{T}\left(\ddfrac{\bz}{\btheta} {}_{\mathrm{right}}D_{C}^\beta \lambda(t)
    -\lambda \frac{\mathrm{d} f}{\mathrm{d} \btheta}\right) \ud t. 
    \end{aligned}
\end{align*}
Since $\bz_{0}$ is the initial input hidden feature of the neural FDE, we have $\frac{\mathrm{d} \bz(0)}{\mathrm{d} \btheta}=0$. Applying the chain rule, we get $\frac{\mathrm{d} f}{\mathrm{d} \btheta}=\frac{\partial f}{\partial \btheta}+\frac{\partial f}{\partial \bz} \ddfrac{\bz}{\btheta}$. Consequently, the derivative of the integral expression is given by:
\begin{align}
\begin{aligned}
& \frac{\mathrm{d}}{\mathrm{d} \btheta}\left[\int_{0}^{T} \lambda F \ud t\right] 
=\lambda\left(T\right) \frac{\mathrm{d} J_{\mathrm{left}}^{1-\beta} \bz(T) }{\mathrm{d} \btheta} \\
&\quad  +\int_{0}^{T}\left({}_{\mathrm{right}}D_{C}^\beta \lambda(t) -\lambda(t) \frac{\partial f}{\partial \bz}\right) \ddfrac{\bz}{\btheta} \ud t  
-\int_{0}^{T} \lambda \frac{\partial f}{\partial \btheta} \ud t.
\end{aligned}
\end{align}
It follows that 
\begin{align}
&\frac{\mathrm{d} L}{\mathrm{d} \btheta}
=\frac{\ud L}{\ud J_{\mathrm{left}}^{1-\beta} \bz(T)} \frac{\mathrm{d} J_{\mathrm{left}}^{1-\beta} \bz(T)}{\mathrm{d} \btheta}
-\frac{\mathrm{d}}{\mathrm{d} \btheta}\left[\int_{0}^{T} \lambda F \ud t\right]\nn
&=\frac{\ud L}{\ud J_{\mathrm{left}}^{1-\beta} \bz(T)} \frac{\mathrm{d} J_{\mathrm{left}}^{1-\beta} \bz(T)}{\mathrm{d} \btheta}
- \lambda\left(T\right) \frac{\mathrm{d} J_{\mathrm{left}}^{1-\beta} \bz(T) }{\mathrm{d} \btheta} \nn
& -\int_{0}^{T}\left({}_{\mathrm{right}}D_{C}^\beta \lambda(t) -\lambda(t) \frac{\partial f}{\partial \bz}\right) \ddfrac{\bz}{\btheta} \ud t  
+\int_{0}^{T} \lambda \frac{\partial f}{\partial \btheta} \ud t \nn
&=\left(\frac{\ud L}{\ud J_{\mathrm{left}}^{1-\beta} \bz(T)} 
- \lambda\left(T\right) \right) \frac{\mathrm{d} J_{\mathrm{left}}^{1-\beta} \bz(T) }{\mathrm{d} \btheta} \nn
& -\int_{0}^{T}\left({}_{\mathrm{right}}D_{C}^\beta \lambda(t) -\lambda(t) \frac{\partial f}{\partial \bz}\right) \ddfrac{\bz}{\btheta} \ud t  
+\int_{0}^{T} \lambda \frac{\partial f}{\partial \btheta} \ud t. \label{eq.supp_fdasfe}
\end{align}

To avoid the direct computation of challenging derivatives such as $\ddfrac{\bz(T)}{\btheta}$ or $\ddfrac{\bz(t)}{\btheta}$, we let $\lambda(t)$ satisfy the following FDE:
\begin{align}
\begin{aligned}
    {}_{\mathrm{right}}D_{C}^\beta \lambda(t) & = \lambda(t) \frac{\partial f}{\partial \bz}, \\
    \text { with } \lambda\left(T\right) &=\frac{\ud L}{\ud J_{\mathrm{left}}^{1-\beta} \bz(T)}.
\end{aligned} \label{eq.supp_back_adjoint}
\end{align}
Consequently, as the first two terms in \cref{eq.supp_fdasfe} vanish, we obtain
\begin{align}
    \frac{\mathrm{d} L}{\mathrm{d} \btheta}=-\int_{T}^{0} \lambda(t) \frac{\partial f}{\partial \btheta} \ud t. \label{eq.supp_back_para}
\end{align}
To facilitate computation, we approximate the constraint on the last time point as
$\frac{\ud L}{\ud J_{\mathrm{left}}^{1-\beta} \bz(T)} \approx \frac{\ud L}{\ud \bz(T)}$ in \cref{eq.supp_back_adjoint}. 
This approximation represents the gradient of the loss with respect to the final state of the system. 
Efficient evaluation of the vector-Jacobian products, $\lambda(t) \frac{\partial f}{\partial \bz}$ and $\lambda(t) \frac{\partial f}{\partial \btheta}$, specified in \cref{eq.supp_back_adjoint} and \cref{eq.supp_back_para}, is achieved using automatic differentiation, offering a computational cost on par with that of evaluating $f$ directly.

In the reverse model, the systems described in \cref{eq.supp_back_adjoint,eq.supp_back_para} are computed simultaneously. Echoing the methodologies employed in \cref{eq.pre,eq.cor}, we utilize standard quadrature techniques to solve these equations. Our numerical iterations confirm that the coefficients are consistent with those reported in \cref{eq.pre,eq.cor}. Additionally, the trajectory $\bz(t)$ generated during the forward pass can be efficiently reused.

Setting $\beta=1$ simplifies the FDE described in \cref{eq.back_adjoint} into a first-order neural ODE. This transformation aligns the computation of the reverse model FDE with the first-order neural ODE framework utilized in \cite{chen2018neural}. It can be formulated in summary as 
\begin{align}
\begin{aligned}
    \frac{\ud }{\ud t} \lambda(t) & = \lambda(t) \frac{\partial f}{\partial \bz}, \\
    \frac{{\mathrm{d} L}/{\mathrm{d} \btheta}}{\ud t} &= \lambda(t) \frac{\partial f}{\partial \btheta},
\end{aligned} \label{eq.supp_back_adjoint_ode}
\end{align}
with $\lambda(T)=\frac{\ud L}{\ud \bz(T)}$ and $\frac{\ud L}{\ud \btheta}(T) = \mathbf{0}_{|\btheta|}$.

\begin{table}[h]
    \centering
    \resizebox{0.5\textwidth}{!}{
\begin{tabular}{ccccccc} 
\toprule
Dataset & Type & Classes & Features & Nodes & Edges \\
\hline Cora & citation & 7 & 1433 & 2485 & 5069  \\
Citeseer & citation & 6 & 3703 & 2120 & 3679\\
PubMed & citation & 3 & 500 & 19717 & 44324 \\
Coauthor CS & co-author & 15 & 6805 & 18333 & 81894 \\
Computers & co-purchase & 10 & 767 & 13381 & 245778 \\
Photos & co-purchase & 8 & 745 & 7487 & 119043  \\
CoauthorPhy & co-author  & 5 & 8415 & 34493 & 247962 \\
OGB-Arxiv & citation & 40 & 128 & 169343 & 1166243 \\
OGB-Products & co-purchase & 47 & 100 & 2449029 & 61859140 \\
\bottomrule
\end{tabular}}
 \caption{Dataset Statistics used in \cref{tab:noderesults,tab:model_products}.}
 \label{tab:nod_dat_sta}
\end{table}
\section{Datasets and Experiments Setting} \label{sec.supp_exp}

\subsection{Graph Datasets Used in the Main Paper}\label{subsec:supp_dataset}
To validate the efficiency of our neural FDE solvers, experiments in \cref{ssec.exp_graph} were performed in a range of graph learning tasks. 
The dataset statistics used in \cref{tab:noderesults,tab:model_products} are provided in \cref{tab:nod_dat_sta}. Following the experimental framework in \citep{chamberlain2021grand}, we select the largest connected component from each dataset.

\begin{table}[H]\small
\centering
\resizebox{0.5\textwidth}{!}{
\begin{tabular}{c|ccccccccc}
    \toprule
    Dataset  & lr & weight decay & indrop & dropout & hidden dim & time & step size  \\
    \midrule
    \multirow{1}{*}{Cora}   & 0.005 & 0.0001 & 0.4 & 0.2 &80  &  4 & 0.2 \\
    \multirow{1}{*}{Citeseer}  & 0.001 & 0.0001 & 0.4 & 0.4 & 64 & 4  & 1\\
    \multirow{1}{*}{PubMed}  & 0.001 & 0.0001  & 0.2 & 0.4 & 64 & 10  & 0.2\\
    \multirow{1}{*}{Coauthor CS}   & 0.005 & 0.0001  & 0.4 & 0.4 & 8 & 8  &1  \\
    \multirow{1}{*}{Computers} & 0.005 & 0.0001 & 0.2 & 0.4 & 64 & 3  &0.5 \\
     \multirow{1}{*}{Photos}   & 0.001 & 0.0001 &  0.2& 0.4 &128  &4  &0.2  \\
     \multirow{1}{*}{CoauthorPhy}   & 0.005 & 0.0001 &  0.2& 0.4 &64  &4  &0.5  \\
    \multirow{1}{*}{OGB-Arxiv}   & 0.005 & 0.0001 &  0.4& 0.2 & 128  &8  &0.2  \\
    \multirow{1}{*}{OGB-Products}   & 0.001 & 0.0001 &  0.2& 0.4 &128  &10  &0.2  \\
    \bottomrule
\end{tabular}}
\caption{Hyper-parameters used in \cref{tab:noderesults}}
\label{tab:s3hyperpara}
\end{table}

\subsection{Hyper-parameters}
We utilized a grid search approach on the validation dataset to fine-tune common hyperparameters including hidden dimensions, learning rate, weight decay, and dropout rate.
The specific hyperparameters utilized in \cref{tab:noderesults} are detailed in \cref{tab:s3hyperpara}. For a comprehensive understanding of the hyperparameter configurations, we direct readers to the accompanying codebase in the supplementary material, which includes the provided code for reproducibility.

\begin{table}[H]
    \small
    \centering 
    
    \setlength{\tabcolsep}{2pt} 
    \resizebox{0.48\textwidth}{!}{
\begin{tabular}{c|c|c|c|c|c|c|c|c|c|c}
    \toprule
$T$          & 1  & 10   & 20   & 30   & 40  & 50    & 60   & 70 & 100 & 170 \\
\hline
{Direct}  & 1600 & 2692 & 3906 & 5122 & 6336 & 7532 & 8746 & 9962 & 13606 & 22072  \\

{Adjoint} & 1552 & 1818 & 2166 & 2510 & 2854 & 3178 & 3526 & 3870 & 4906 & 7298 \\
$\frac{\text{Adjoint}}{\text{Direct}}$ & 0.97 & 0.68 & 0.55 & 0.49 & 0.45 & 0.42 & 0.40 & 0.39 & 0.36 & 0.33 \\
    \bottomrule
    \end{tabular}
    }
    \caption{GPU memory usage (MB) during training on the Fashion-MNIST dataset across various integral times $T$ with $h=0.1$. }
    \label{tab:fmnist_T}
\end{table}

\subsection{More Experiments on Image Dataset} 
This supplementary document extends the adjoint backpropagation evaluation of neural FDEs on the MNIST dataset \cite{lecun1998gradient}, initially presented in \cref{ssec.image} of the main paper, to the Fashion-MNIST dataset \cite{xiao2017fashion}. Fashion-MNIST consists of 28$\times$28 grayscale images of 70,000 fashion items across 10 categories, with each category containing 7,000 images. It mirrors MNIST in image size, data format, and the structure of training and testing splits, featuring 60,000 images in the training set and 10,000 in the test set. In our experiments, the model's hidden features are updated using a neural FDE where $f(t,\bz(t);\btheta)$ is implemented as a convolution module. The step size $h$ is set at 0.1, and the fractional order $\beta$ at 0.3. Batch sizes of 128 are used during both the training and testing phases.
In the first setting, the gradients are backpropagated directly through forward-pass operations in \cref{eq.pre}, while in the second setting, we solve the proposed reverse-mode FDE in \cref{ssec.reverse_fde} to obtain the gradients.
When $T=1$, we find training with the adjoint backpropagation achieves a test accuracy of over 92.05\% while training with the direct differentiation achieves 92.35\%. 
Furthermore, we set integral times $T$ ranging from 1 to 170 and record the GPU memory usage in \cref{tab:fmnist_T}. 
We observe that with sufficiently large $T$, the adjoint backpropagation consumes only 33\% of the training memory compared to the direct differentiation.
Additionally, we observe that the GPU memory during training scales linearly with the integration time $T$ with a fixed step size $h$. This meets our complexity analysis in \cref{ssec.complex}.

\section{Limitations and Broader Impact} \label{sec.supp_limit}

While the proposed scalable adjoint backpropagation method for training neural FDEs introduces significant improvements in memory efficiency and computational overhead, there are several limitations that merit attention. 
For instance, this method may introduce numerical inaccuracies due to interpolation errors and sensitivity to initial conditions when solving the augmented FDE backward in time. This sensitivity could impact the robustness and generalizability of the models trained using this method, particularly in chaotic FDE systems.
The integration of fractional-order calculus with deep learning opens new possibilities for modeling complex dynamical systems with greater accuracy and flexibility than traditional integer-order models. This advancement has broad implications across various fields. For example, enhanced modeling capabilities could lead to more precise simulations and predictions in physics, biology, and engineering, facilitating breakthroughs in understanding complex systems. Additionally, by reducing computational requirements, smaller organizations and researchers with limited resources may also leverage advanced models, potentially democratizing access to cutting-edge AI tools.

\end{document}